\documentclass[journal]{IEEEtran}

\usepackage[hyphens]{url}
\usepackage{graphicx}
\usepackage{amsmath}
\usepackage{amssymb}
\usepackage{booktabs}
\usepackage{makecell}

\begin{document}

\title{Differential Unfolding: 
Efficient Reconstruction for
Video Snapshot Compressive Imaging}

\author{Yin-Ping Zhao,~\IEEEmembership{Member,~IEEE,}
Muyuan Zhang, 
Jiaxi Li, \\
Jiancheng Zhang, 
Haijin Zeng,~\IEEEmembership{Member,~IEEE}
\IEEEcompsocitemizethanks{
    \IEEEcompsocthanksitem This work was supported in part by the National Natural Science Foundation of China under Grants 62302394 and 62671510, and in part by the China Postdoctoral Science Foundation under Grant 317751.
    \IEEEcompsocthanksitem Yin-Ping Zhao and Jiaxi Li are with the School of Cyberspace Security, Northwestern Polytechnical University, Xi'an 710072, China.
   \IEEEcompsocthanksitem Muyuan Zhang is with the School of Software, Northwestern Polytechnical University, Xi'an 710072, China.
    \IEEEcompsocthanksitem Jiancheng Zhang is with the School of Computer Science and Technology, Hong Kong University, Hong Kong, China.
    \IEEEcompsocthanksitem Haijin Zeng is with the School of Computer Science and Technology, Harbin Institute of Technology, Shenzhen, 518055, China. (Corresponding author: zenghj@hit.edu.cn)
    }
}%

\maketitle

\begin{abstract}
Although Deep Unfolding Networks (DUNs) are predominant in Video Snapshot Compressive Imaging (SCI), they are still limited by a uniform design philosophy. The existing approaches stack high-complexity priors with the same structure repeatedly, neglecting the fact that the optimization paths gradually converge to a nearly static state. As a consequence, representation stagnation occurs, and much computational cost is wasted on small feature updates. To solve this inefficiency, we propose Differential Unfolding (DU), a heterogeneous framework that substitutes repeated work with a periodic schedule of heavy and lightweight stages.
The core of DU is the Differential Evolutionary Framework (DEF), which divides the unfolding procedure into two complementary roles: structural anchoring and differential evolution. In this method, high-parameter general stages are used sparsely to form high-fidelity feature foundations. On the other hand, lightweight differential stages use a Differential Representation Prior (DRP) to pass on and refine these foundational features by a differential mechanism. By integrating Differential Representation Attention (DRA) for evolving attention maps and a Differential Modulated FFN (DM-FFN) for feature rectification, DRP effectively models cross-stage variations with minimal overhead. By concentrating high-capacity computation on the sparsely-scheduled general stages and delegating incremental refinement to the lightweight differential stages, DU achieves a superior trade-off between accuracy and efficiency. Extensive experiments show that DU-9stg achieves state-of-the-art reconstruction quality, while DU-5stg surpasses all compared methods with the fewest parameters and FLOPs; compared with the unfolding method DADUN, DU-9stg needs about half the parameters and a quarter of the FLOPs.
The code is at https://github.com/Muyuan-Zhang/DU.
\end{abstract}


\IEEEpeerreviewmaketitle

\section{Introduction}
\label{sec:intro}
\IEEEPARstart{V}{ideo} Snapshot Compressive Imaging (SCI)~\cite{cao2026sparse,yin2025wavelet,zhong2024untrained,videosci1,videosci2,videosci3,videosci4,videosci5,videosci6}, rooted in the principles of Compressive Sensing (CS)~\cite{cs}, has enabled the acquisition of high-speed video with low-speed sensors. It encodes a sequence of high-speed frames into a single 2D measurement captured by a conventional camera. The same encoding principle has also been extended to other modalities, such as spectral imaging~\cite{led-sci,hbn-cassi,cassi-progressive,cassi-complementary} and coherence tomography~\cite{sct-imaging}.
However, the subsequent recovery of the original video constitutes a severely ill-posed inverse problem.
A spectrum of solutions has been proposed to tackle this challenge, ranging from traditional model-based~\cite{gap-tv,desci,gmm,gmm2}
\begin{figure}[t]
    \centering
    \includegraphics[width=0.98\columnwidth]{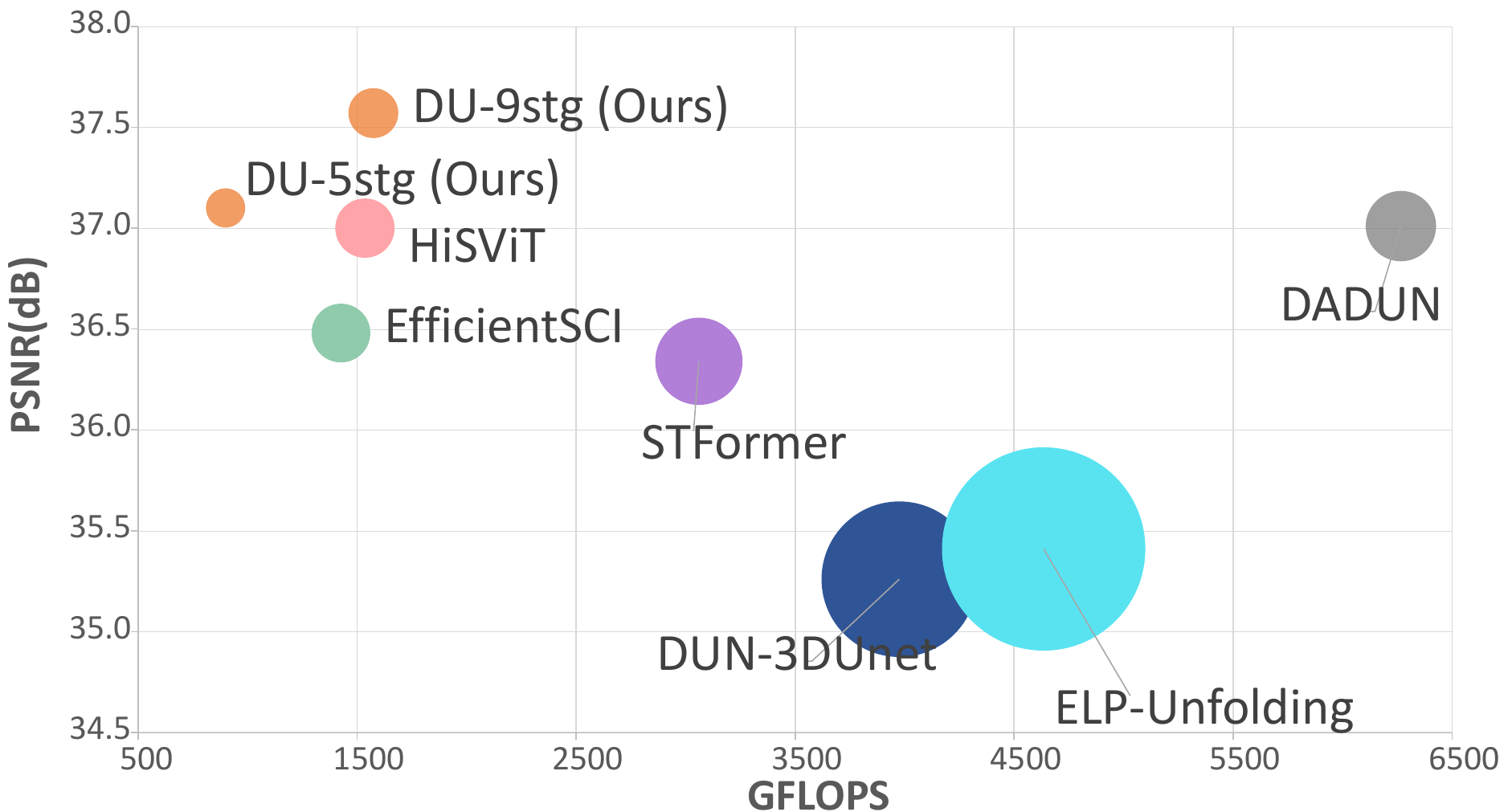}
    \caption{Comparison of PSNR-FLOPs-Params, averaged over the 6 grayscale benchmark datasets (Kobe, Traffic, Runner, Drop, Crash, Aerial), between the proposed DU-5stg/DU-9stg (Ours) and competing deep unfolding methods (DUN-3DUnet~\cite{dun-3dunet}, ELP-Unfolding~\cite{elp}, DADUN~\cite{dadun}) and end-to-end Transformer-based methods (STFormer~\cite{stformer}, EfficientSCI~\cite{efficientsci}, HiSViT~\cite{hisvit}). The radius of each marker indicates the number of parameters.}
    \label{fig:psnr}
\end{figure}
and plug-and-play~\cite{pnp-ffdnet,pnp-fastdcdnet,cqnpm-mri,gksm-mri} methods to modern end-to-end learning approaches~\cite{e2e-cnn,revsci,birnat,metasci,stformer,efficientsci,deepoptics,hisvit,madygraph,mobilesci,qsci}.
Among these, deep unfolding methods~\cite{admm-net,ctm-sci,dun-3dunet,elp,gap-net,step-sci,spadun,dadun} have emerged as a promising direction. These deep unfolding networks (DUNs) unroll an iterative optimization algorithm into network stages, combining the structure of classical algorithms with the learning capacity of deep networks.
\begin{figure*}[!t]
    \centering
    \includegraphics[width=.98\linewidth]{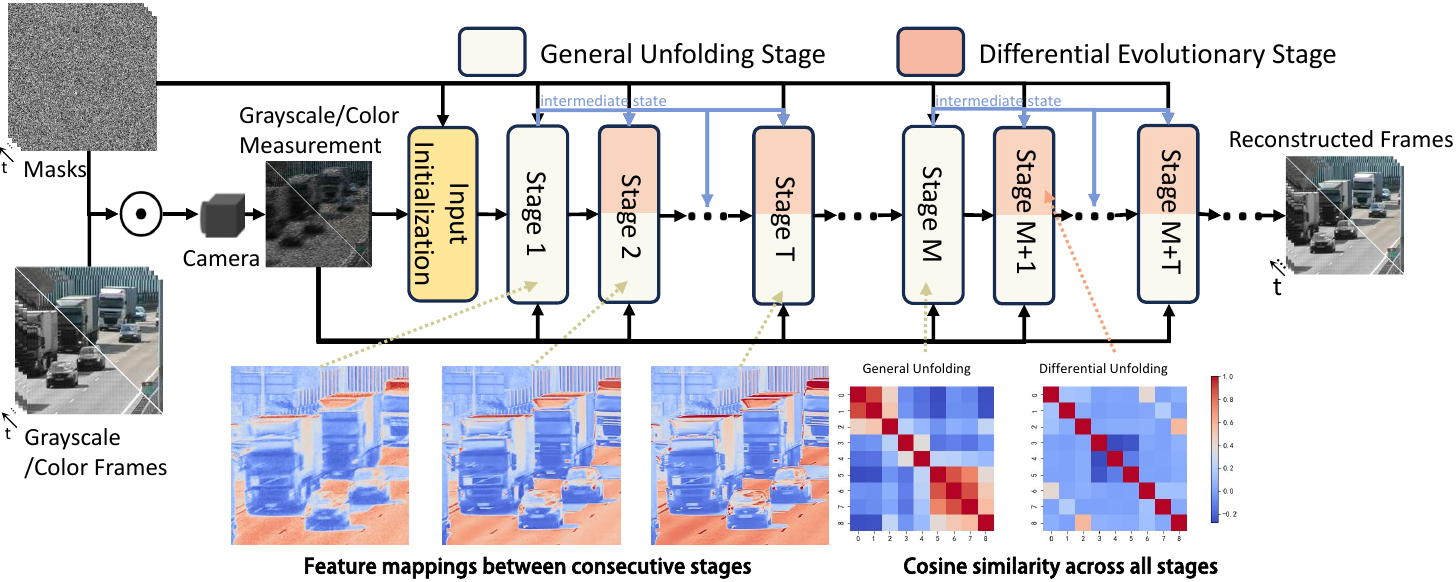}
    \caption{
    Overview of the video SCI forward model and the proposed unfolding-based reconstruction pipeline. Masked, temporally integrated frames are recovered by stages that periodically alternate between general unfolding (cream) and differential evolutionary (salmon) stages, with the intermediate state passed across stages (blue arrows). The near-identical feature heatmaps from consecutive general-unfolding stages (bottom left) illustrate inter-stage representation homogeneity: repeatedly applying a full-capacity prior yields only marginal representation change. The cosine-similarity matrices (bottom right) quantify this---uniformly high (red) under a homogeneous baseline versus substantially lower (blue) under our differential unfolding---confirming that the differential stages keep contributing non-redundant updates.
    }
  \label{fig:UF}
\end{figure*}
Although they are successful, most of the current DUNs have the same structural limitation, namely a homogeneous stage design in which prior networks with identical architectures and comparable capacity are repeatedly used across stages. 
Although this approach preserves interpretability and maintains a rigorous correspondence with the iterative optimization procedure, it neglects the inherent dynamic characteristics of the reconstruction process. 
During execution, the tendency of the reconstruction to converge leads to smaller changes in intermediate feature states.
However, the traditional prior modules still perform the full prior transformation at each stage, regardless of the reduced magnitude of state updates.
This architectural mismatch between saturating representation evolution and static model leads to what we term inter-stage representation homogeneity and introduces computational redundancy.
From our experimental results, this phenomenon can be confirmed. As shown in Fig.~\ref{fig:UF}, the feature heatmaps of consecutive stages are highly similar, indicating that the variation of representation among stages is small. The cosine similarity between the features of different stages is still high, suggesting that the successive updates bring only slight modifications. Under these circumstances, the repeated use of an entire prior network does not greatly change the representation but strengthens the previously formed structures.
Beyond the qualitative feature heatmaps in Fig.~\ref{fig:UF}, we further observe how the inter-stage feature update evolves along the unfolding trajectory. At first, both the homogeneous baseline and DU exhibit obvious updates. However, in the baseline, the update becomes weaker in the later stages, indicating that its uniformly heavy prior networks perform largely redundant operations when the reconstruction tends to converge. In contrast, the update in DU decays more slowly, showing that its differential stages continuously improve the reconstruction without causing unnecessary computations.
These findings indicate a structural limitation of the existing DUNs: the lack of a mechanism to distinguish between full representation reconstruction and incremental evolution.

To address the inefficiency caused by inter-stage representation homogeneity, we propose Differential Unfolding (DU), a novel paradigm for efficient video SCI reconstruction.
Unlike conventional unfolding designs that repeatedly apply identical priors throughout the optimization trajectory, DU introduces structural heterogeneity into the unfolding process via a periodic schedule, explicitly separating stages that reconstruct representations from scratch from stages that refine them incrementally.
The proposed framework is built upon two key components: the Differential Evolutionary Framework (DEF) and the Differential Representation Prior (DRP).
Departing from homogeneous architectures, DEF partitions the reconstruction trajectory into periodic general unfolding stages and differential evolutionary stages. General stages employ a high-parameter spatial-temporal backbone for holistic restructuring, while differential stages focus on incremental updates as inter-stage variations become increasingly subtle.
This periodic alternation is motivated by the non-uniform dynamics of iterative reconstruction. By reintroducing general unfolding, DEF prevents error accumulation and representation drift, while intermediate refinement stages circumvent redundant computation on nearly converged states.
To efficiently model incremental evolution, the DRP captures cross-stage variations rather than reconstructing features from scratch. It consists of Differential Representation Attention (DRA), which tracks evolving attention dependencies, and a Differential Modulated Feed-Forward Network (DM-FFN), which adaptively modulates features via differential-driven gating.
This stage-wise design reduces computational redundancy associated with inter-stage representation homogeneity while preserving essential structural updates, thereby improving efficiency and the modeling of dynamic regions and fine details.

Our main contributions are summarized as follows:
\begin{itemize}

\item We identify and systematically analyze inter-stage representation homogeneity in deep unfolding networks, revealing it as a fundamental source of computational redundancy during iterative reconstruction.

\item We propose the Differential Unfolding (DU) method with a Differential Evolutionary Framework (DEF), which introduces a periodic unfolding mechanism that alternates between general unfolding and differential evolutionary stages on a fixed schedule.

\item We design a Differential Representation Prior (DRP) that explicitly models cross-stage representation evolution, enabling efficient refinement under highly similar feature states and reducing unnecessary computation.

\item Extensive experiments demonstrate that DU achieves state-of-the-art reconstruction performance while significantly reducing parameter count and computational complexity across multiple SCI benchmarks.
\end{itemize}

\section{Related Work}
\subsection{Video SCI Reconstruction}
Traditional model-based algorithms~\cite{gap-tv,desci} offer rigorous mathematical interpretability, but they have gradually been abandoned because of their long solving times and low reconstruction quality.
With the development of deep learning, deep neural networks have become the focus of research owing to their faster reconstruction speed and higher reconstruction quality. For instance, BIRNAT~\cite{birnat} employs bidirectional recurrent neural networks for reconstruction.
EfficientSCI~\cite{efficientsci} integrates the local modeling ability of convolutional networks with the long-range modeling ability of transformers, achieving a reduction in the computational cost of the model through dense connections.
HiSViT~\cite{hisvit} considers the imaging characteristics of video SCI systems and emphasizes spatial restoration in the early stages of the model.
Beyond supervised training, unsupervised reconstruction has also been explored: Miao et al.~\cite{dvp-sci} propose a domain-factorized deep video prior that reconstructs video SCI measurements at test time via a tailored network prior and affine motion compensation, without requiring any training data.

\subsection{Deep Unfolding Methods}
Previous model-based methods typically require manually designed priors and lack flexibility when handling different reconstruction scenarios.
The deep unfolding method~\cite{admm-net,ctm-sci,dun-3dunet,elp,gap-net,step-sci} combines traditional mathematical iterative algorithms with deep neural networks, retaining both mathematical interpretability and the ability to adaptively adjust the priors.
ELP-Unfolding~\cite{elp} leverages ensemble priors to achieve both high reconstruction speed and accuracy within a single network. DUN-3DUnet~\cite{dun-3dunet} considers information loss between stages and uses DFMA to fuse information across stages. Outside video SCI, several unfolding networks also reuse or filter information across stages: MADUN~\cite{madun} maintains a persistent inter-stage memory with self-modulation, and GCDUN~\cite{gcdun} uses LSTM-inspired gates to filter redundant information before each proximal gradient step. Other unrolled networks for MRI couple image- and edge-specific regularizers within each stage~\cite{mri-edge-unfold} or parameterize an unrolled nonlocal network via learned group sparsity~\cite{groupcdl}. Our DEF differs in that it structurally alternates high-capacity and lightweight stages on a fixed periodic schedule, rather than augmenting every stage with the same memory or gating mechanism.
DADUN~\cite{dadun} introduces a degradation-aware deep unfolding framework that uses estimated priors and a BiP-CRNN to guide iterative reconstruction of video SCI sequences.
Cross-stage information reuse and feature-difference-driven modulation have each been explored individually (e.g., DUN-3DUnet's inter-stage fusion, MADUN's persistent memory, and attention-difference or residual-update mechanisms outside video SCI). DU instead introduces a periodic heavy/light unfolding schedule together with the DRA/DM-FFN differential operators that instantiate it for video SCI.
\section{Proposed Method}

\subsection{Mathematical Model of Video SCI}
In the video SCI system, three-dimensional high-speed video frames are modulated by multiple masks, then integrated along the temporal dimension, and finally captured by a two-dimensional camera sensor as compressed two-dimensional measurements.
Fig.~\ref{fig:UF} illustrates the compression process of video SCI.
Mathematically, high-speed grayscale video frames $F\in\mathbb{R}^{H\times W\times T}$ are modulated by masks with patterns $M\in \mathbb{R}^{H\times W \times T}$. After temporal compression, the two-dimensional compressed measurement $G\in \mathbb{R}^{H\times W}$ is captured by the camera sensor. This process can be expressed as:
\begin{equation}
G(h,w)=\sum_{i=1}^{T}M(h,w,i)\odot F(h,w,i)+N,\label{eq1}
\end{equation}
where $\odot$ denotes the element-wise multiplication (Hadamard product), $h$ and $w$ represent spatial coordinates, and $N\in \mathbb{R}^{H\times W}$ represents noise introduced during the imaging process.
The matrix-vector form of Eq.~\eqref{eq1} can be written as:
\begin{equation}
g=\mathrm{\Phi} f+n,\label{eq2}
\end{equation}
where $g=\text{vec}(G)\in \mathbb{R}^{HW}$, $f=\text{vec}(F)\in \mathbb{R}^{HWT}$, and $n=\text{vec}(N)\in\mathbb{R}^{HW}$ denote the vectorized forms of the measurement $G$, the original video frames $F$, and the noise $N$, respectively, and $\mathrm{\Phi}=[\mathrm{D}_1,\mathrm{D}_2,\ldots,\mathrm{D}_T]\in\mathbb{R}^{HW\times HWT}$ is the sensing matrix built from the masks, with $\mathrm{D}_i=\mathrm{Diag}(\text{vec}(M(:,:,i)))\in\mathbb{R}^{HW\times HW}$ a diagonal matrix whose diagonal entries are filled by the vectorized $i$-th mask~\cite{stformer}.
\subsection{Extension to Color Video SCI}
\label{sec:color-math}
For a color video SCI system, each pixel captures only the red (R), green (G), or blue (B) channel of the raw data in a spatial Bayer layout. The input color video frames $F\in\mathbb{R}^{H\times W\times T\times3}$ and mask $M\in\mathbb{R}^{H\times W\times T}$ are divided into four sub-channels $\{F^r,F^{g_1},F^{g_2},F^{b}\}\in\mathbb{R}^{\frac{H}{2}\times\frac{W}{2}\times T}$ and $\{M^r,M^{g_1},M^{g_2},M^b\}\in\mathbb{R}^{\frac{H}{2}\times\frac{W}{2}\times T}$, respectively. The four sub-measurements are obtained as:
\begin{align}
G^r(h,w)&=\sum_{i=1}^{T}M^r(h,w,i)\odot F^r(h,w,i)+N^r,\\
G^{g_1}(h,w)&=\sum_{i=1}^{T}M^{g_1}(h,w,i)\odot F^{g_1}(h,w,i)+N^{g_1},\\
G^{g_2}(h,w)&=\sum_{i=1}^{T}M^{g_2}(h,w,i)\odot F^{g_2}(h,w,i)+N^{g_2},\\
G^{b}(h,w)&=\sum_{i=1}^{T}M^{b}(h,w,i)\odot F^{b}(h,w,i)+N^{b}.
\end{align}
Following previous methods~\cite{stformer,efficientsci}, we simultaneously feed the four sub-measurements into the network to reconstruct the RGB color video directly, without relying on separate demosaicing algorithms.

\subsection{General Unfolding Methods}
Traditional deep unfolding algorithms build the unfolding framework by solving the following subproblem:
\begin{equation}
\hat{f}=\underset{f}{\mathrm{argmin}}\frac{1}{2}||g-\Phi f||^2+\tau R(f), \label{eq3}
\end{equation}
where $\frac{1}{2}||g-\Phi f||^2$ is the data fidelity term, $R(f)$ is the image prior term, and $\tau$ is the hyperparameter that balances their importance.
For example, in the case of the HQS method, the problem is then decomposed into two alternating steps: a gradient descent step and a prior mapping step, which are solved iteratively as follows:
\begin{gather}
z_{k} = f_{k-1} + \rho \mathrm{\Phi}^\mathsf{T} \left[  \left( g - \mathrm{\Phi} f_{k-1} \right)  ./ \left( \mathrm{\Phi \Phi^\mathsf{T}} \right)\right], \label{eq4} \\
f_{k} = \mathcal{P}_{k}(z_{k}),
\label{eq5}
\end{gather}
where $k$ represents the iteration stage, $\rho$ is the step size for the iteration, which can be estimated by a simple network from~\cite{dauhst}, $\mathcal{P}$ denotes the prior mapping module, and $./$ denotes elementwise division against the diagonal of $\Phi\Phi^\mathsf{T}$.
Fig.~\ref{fig:UF} depicts the overall unfolding pipeline, and Fig.~\ref{fig:PU}(a) shows the structure of a single general stage.
We follow previous works~\cite{birnat,dun-3dunet,efficientsci} and use a similar input initialization module to obtain a good initial input $f_0$.
While this formulation is effective, it relies on a homogeneous structure, where the same prior network architecture is used across all stages of the unfolding process.
This leads to the inter-stage representation homogeneity discussed in Section~\ref{sec:intro}, where successive stages become increasingly similar, causing computational redundancy and inefficiencies. To address this issue, we introduce the Differential Evolutionary Framework (DEF), which restructures the unfolding process by interleaving general unfolding and differential evolutionary stages on a fixed periodicity $P$. The period $P$ is a fixed hyperparameter, so DEF is a periodic heterogeneous schedule rather than an input-adaptive one; its effect, including removing the periodic general stages entirely, is studied in Fig.~\ref{fig:periodicity}.

\subsection{Differential Evolutionary Framework}
Starting from the general unfolding framework, we decompose the prior mapping module $\mathcal{P}_k$ in Eq.(\ref{eq5})
into an encoder $\mathcal{E}_k$ and a decoder $\mathcal{D}_k$, and obtain the intermediate state $s$ as the reference feature. This can be mathematically represented as:
\begin{align}
s_k &= \mathcal{E}_k(z_{k}), \\
f_{k} &= \mathcal{D}_{k}(s_k).
\end{align}
The intermediate state $s_k$ contains rich feature representations accumulated during the $k$-th unfolding iteration stage, which is then used in the next stage for refinement modeling reference, enabling the update of redundant components while simultaneously generating a new reference intermediate state $s_{k+1}$.
To balance the recovery of primary features with the update of redundant information, the unfolding stages are divided into general unfolding stages, responsible for primary feature reconstruction, and differential evolutionary stages, responsible for refinement updates.
Mathematically, the unfolding process can be described as follows:
\begin{figure}[!t]
    \centering
    \includegraphics[width=\columnwidth]{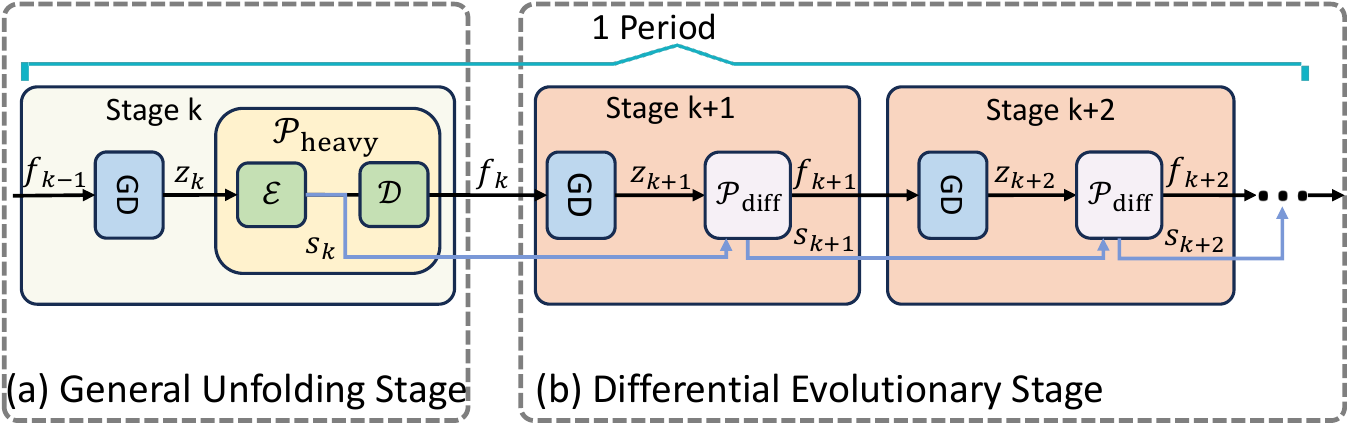}
    \caption{
    Illustration of the Differential Evolutionary Framework (DEF), which alternates general unfolding and differential evolutionary stages on a fixed period. (a) A general unfolding stage applies a gradient-descent (GD) step followed by a high-parameter prior $\mathcal{P}^{\text{heavy}}$ (encoder $\mathcal{E}$, decoder $\mathcal{D}$) to produce $f_k$ and an intermediate state $s_k$. (b) Each subsequent differential evolutionary stage performs GD, then refines the result with a lightweight $\mathcal{P}^{\text{diff}}$ conditioned on the previous stage's intermediate state, producing both the output and a new intermediate state passed forward within the period. A single general stage thus anchors an entire period of lightweight differential refinements.
    }
  \label{fig:PU}
\end{figure}
\begin{equation}
f_{k+1} =
\begin{cases}
\mathcal{P}^{\text{heavy}}_{k+1}\big(z_{k+1}\big), & \text{General Unfolding Stage},\\[2mm]
\mathcal{P}^{\text{diff}}_{k+1}\big(z_{k+1}, s_k\big), & \text{Differential Evolutionary Stage},
\end{cases}
\label{eq:def}
\end{equation}
where $\mathcal{P}^{\text{heavy}}_{k+1}$ represents a high-parameter prior module designed to drastically model the feature representation, and
$\mathcal{P}^{\text{diff}}_{k+1}$ represents a lightweight differential refinement operation designed to capture the differences between consecutive stages as shown in Fig.~\ref{fig:PU}.
Following the encoder-decoder decomposition $\mathcal{P}_k=\mathcal{D}_k\circ\mathcal{E}_k$ introduced above, each branch of Eq.~(\ref{eq:def}) likewise factorizes into an encoder and a decoder, so applying $\mathcal{P}^{\text{heavy}}_{k+1}$ or $\mathcal{P}^{\text{diff}}_{k+1}$ also yields the next intermediate state $s_{k+1}$ before it is decoded into $f_{k+1}$, matching the $s_{k+1}$ output depicted in Fig.~\ref{fig:PU}.

This periodic design avoids redundant high-capacity computation while enabling targeted incremental updates, improving both efficiency and convergence.

This design also admits a natural interpretation from an optimization standpoint. Unlike standard DUNs that recompute an absolute state estimate $\mathcal{P}(z_k)$ at every stage, DU's differential stages instead learn the residual update of the proximal operator in HQS: conditioning $\mathcal{P}^{\text{diff}}_{k+1}$ on the preceding intermediate state $s_k$ mirrors Nesterov-style momentum extrapolation, using the difference between consecutive states to predict a more accurate search direction, while the sparsely interspersed general stages act as standard proximal mappings that re-anchor the trajectory. This view casts DU's periodic heavy/light schedule as eliminating the redundant re-estimation performed by conventional DUNs at every stage, in favor of a more efficient optimization trajectory.

\begin{figure}[!t]
    \centering
    \includegraphics[width=.98\linewidth]{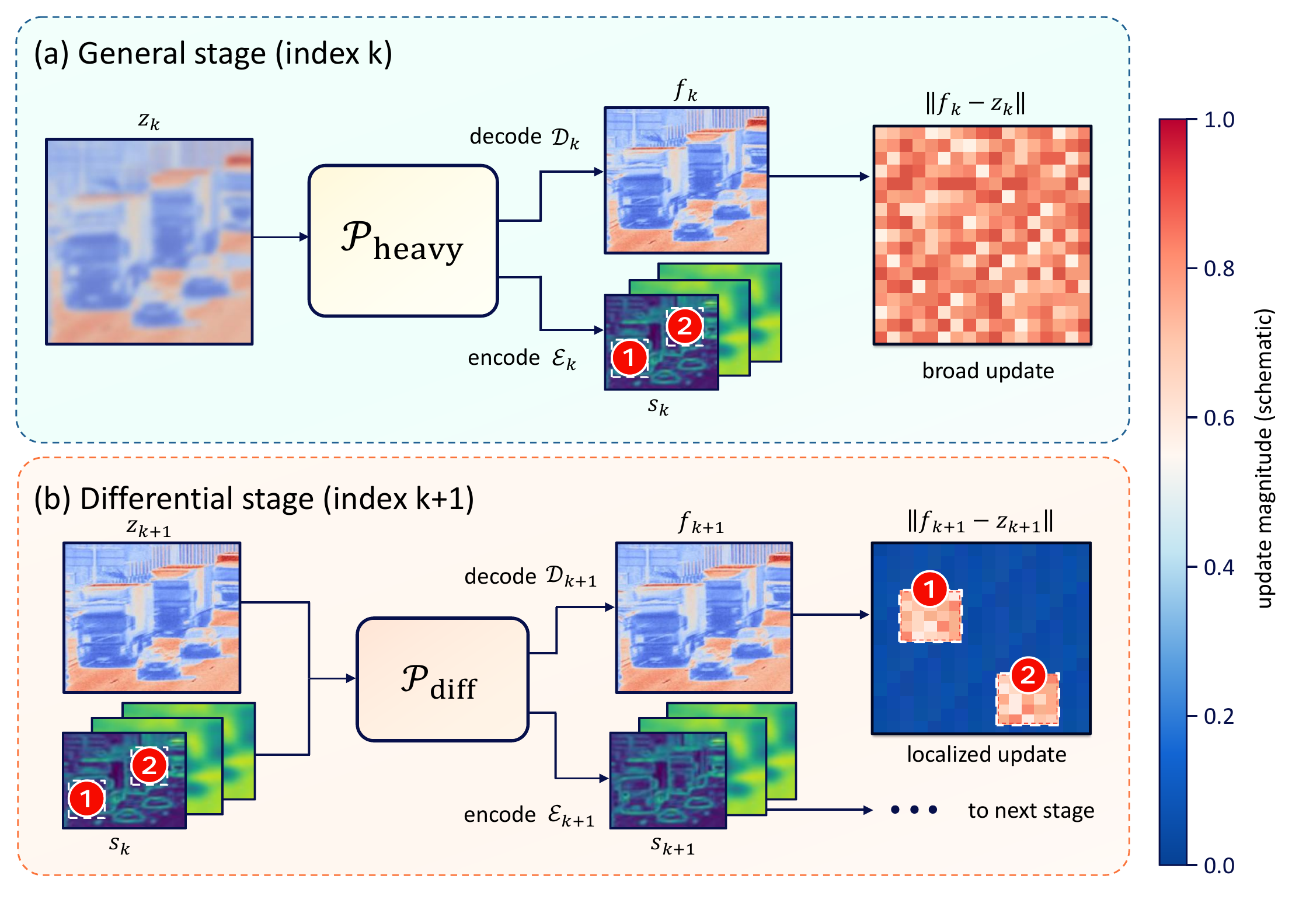}
    
\caption{Schematic illustration of the general and differential unfolding stages in Eq.~\eqref{eq:def}. (a) The general stage takes $z_k$ and produces the reconstruction $f_k$ and intermediate feature state $s_k$, yielding broad changes in the frame-domain update map. (b) The differential stage refines $z_{k+1}$ conditioned on $s_k$, producing $f_{k+1}$ and the next feature state $s_{k+1}$. Its update map is localized around the marked history-guided regions. The difference maps illustrate frame-domain updates, whereas $s_k$ denotes multi-channel encoder features rather than a reconstructed image.}

  \label{fig:gendiff}
\end{figure}

\begin{figure*}[!t]
    \centering
    \includegraphics[width=.98\linewidth]{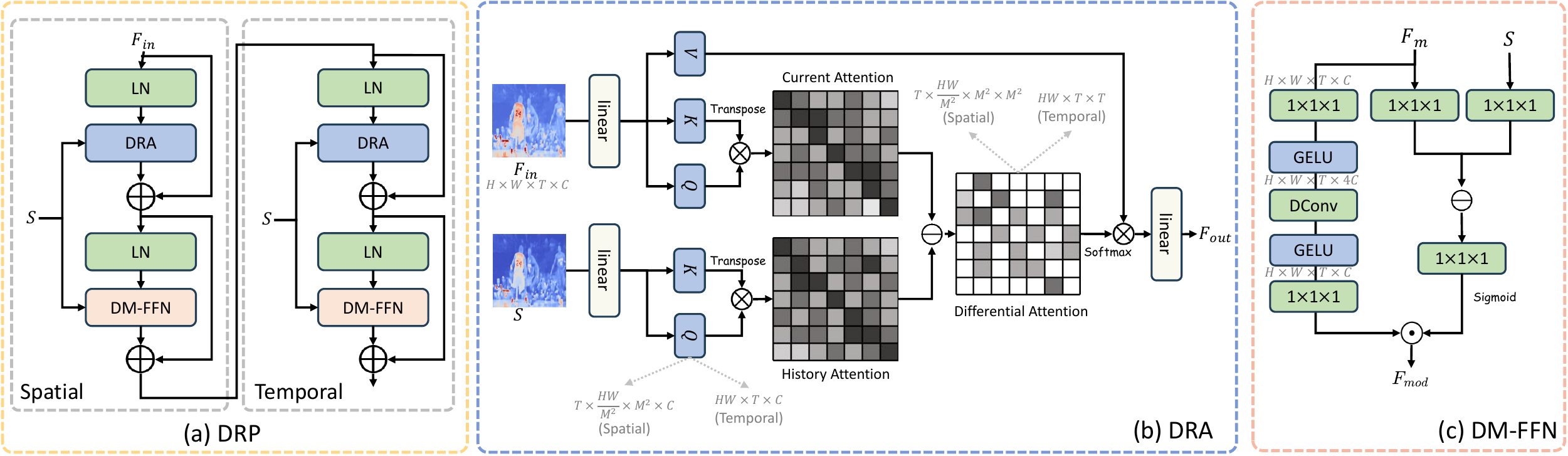}
    \caption{
    Architecture of the Differential Representation Prior (DRP), the lightweight module $\mathcal{P}^{\text{diff}}$ used in differential evolutionary stages. (a) DRP backbone: spatial and temporal branches each apply Differential Representation Attention (DRA) and a Differential Modulated Feed-Forward Network (DM-FFN), with residual connections to the intermediate state $S$. (b) DRA: the difference between attention maps computed from the current feature $F_{in}$ and the intermediate state $S$ yields a differential attention map, so the module attends to regions that changed between stages rather than recomputing a full attention map. (c) DM-FFN: a feature branch extracts local spatiotemporal features, while a modulation branch gates them via the sigmoid-activated difference between $F_m$ and $S$, amplifying changed regions and suppressing converged ones.
    }
  \label{fig:network}
\end{figure*}

Fig.~\ref{fig:gendiff} further illustrates the contrasting update patterns of the two stages, following the indexing of Eq.~\eqref{eq:def}. The general stage at index $k$ takes the current state $z_k$ and produces both the decoded frame $f_k=\mathcal{D}_k(s_k)$ and the intermediate state $s_k=\mathcal{E}_k(z_k)$. Since $z_k$ and $f_k$ share the same frame-domain shape (Eqs.~\eqref{eq4} and \eqref{eq5}), the update magnitude is schematically visualized as $\|f_k-z_k\|$, which is broadly distributed, indicating a substantial update across the reconstruction. The intermediate state $s_k$, as an encoder feature representation, is depicted as a stack of feature maps. The differential stage at index $k+1$ is conditioned on two inputs, as in Eq.~\eqref{eq:def}: the current state $z_{k+1}$ and the previous stage's intermediate state $s_k$. The informative regions in $s_k$ (marked 1 and 2) schematically correspond to the localized updates in $\|f_{k+1}-z_{k+1}\|$, where $f_{k+1}=\mathcal{P}^{\mathrm{diff}}_{k+1}(z_{k+1},s_k)$, while the remaining regions undergo relatively minor changes. Meanwhile, a new intermediate state $s_{k+1}$ is generated for subsequent refinement stages. This selective, history-guided behavior is realized by the proposed DRP through differential attention in DRA and feature modulation in DM-FFN, as detailed in Fig.~\ref{fig:network}.

To efficiently implement $\mathcal{P}^{\mathrm{diff}}$ in the evolutionary refinement stage, the proposed DRP captures differential variations between consecutive stages, ensuring that meaningful updates are applied while redundant computation is minimized.

\subsection{Differential Representation Prior}
The DRP consists of two core components: Differential Representation Attention (DRA), which effectively updates features by computing the differences between the attention maps of the current feature representation and the intermediate state from the previous stage; and Differential Modulated Feed-Forward Network (DM-FFN), which modulates the spatial-temporal enhanced features by generating modulation parameters through a difference-driven convolutional module.

\subsubsection{Network Architecture}

As demonstrated in Fig.~\ref{fig:network} (a), DRP contains two cascaded spatial-temporal factorized transformer layers~\cite{vivit,timesformer}, which employ DRA to update information in the spatial and temporal domains, respectively.
Following DRA, the proposed DM-FFN leverages the intermediate state to generate modulation information for feature rectification.
Four Layer Normalizations (LN) are employed to stabilize training and enhance generalization ability.

\subsubsection{Differential Representation Attention}
DRA computes differential attention by subtracting the similarity matrix of the previous intermediate state from that of the current feature: correlations that the current feature strengthens relative to the previous state are emphasized, while those unchanged between consecutive stages are suppressed toward uniform weights, focusing computation on the evolving aspects of the reconstruction.
As illustrated in Fig.~\ref{fig:network} (b), the input feature $F_{in}\in\mathbb{R}^{H\times W\times T\times C}$ embedded by an implicit position~\cite{cpe} is linearly projected into the query
$Q_{in}\in\mathbb{R}^{H\times W\times T\times C}$, key $K_{in}\in\mathbb{R}^{H\times W\times T\times C}$ and value $V_{in}\in\mathbb{R}^{H\times W\times T\times C}$ as:
\begin{equation}
Q_{in}=F_{in}W_{Q_{in}},\quad K_{in}=F_{in}W_{K_{in}},\quad V_{in}=F_{in}W_{V_{in}},
\end{equation}
where $W_{Q_{in}}\in\mathbb{R}^{C\times C}$,
$W_{K_{in}}\in\mathbb{R}^{C\times C}$ and
$W_{V_{in}}\in\mathbb{R}^{C\times C}$ are learnable parameters.
The intermediate state $S\in\mathbb{R}^{H\times W\times T\times C}$ is linearly projected into query
$Q_{r}\in\mathbb{R}^{H\times W\times T\times C}$ and key $K_{r}\in\mathbb{R}^{H\times W\times T\times C}$ as:
\begin{equation}
Q_{r}=SW_{Q_{r}},\quad K_{r}=SW_{K_{r}},
\end{equation}
where $W_{Q_{r}}\in\mathbb{R}^{C\times C}$ and $W_{K_{r}}\in\mathbb{R}^{C\times C}$ are learnable parameters.
For the spatial domain, $Q_{in},K_{in},V_{in},Q_{r},K_{r}$ are partitioned into non-overlapping windows of $m\times m$ tokens and reshaped into $\mathbb{R}^{T\times\frac{HW}{m^2}\times m^2\times C}$; for the temporal domain, $Q_{in},K_{in},V_{in},Q_{r},K_{r}$ are reshaped into $\mathbb{R}^{H W\times T\times C}$.
$Q_{in},K_{in},V_{in},Q_{r},K_{r}$ are subsequently split along the channel dimension into $n_h$ attention heads.
The dimension of each head is $d=\frac{C}{n_h}$.
The illustration in Fig.~\ref{fig:network} (b) depicts the situation where $n_h=1$ and some details are omitted for simplification.
The similarity matrix for each head for $F_{in}\in\mathbb{R}^{H\times W\times T\times C}$ and $S\in\mathbb{R}^{H\times W\times T\times C}$ are computed as:
\begin{equation}
\begin{aligned}
A_{in}&=Q_{in}K_{in}^T,\\
A_{r}&=Q_{r}K_{r}^T.
    \label{eq14}
\end{aligned}
\end{equation}
The spatial similarity matrix is given by $A_{in},A_{r}\in\mathbb{R}^{ T\times\frac{HW}{m^2}\times m^2\times m^2}$, and the temporal similarity matrix is given by $A_{in},A_{r}\in\mathbb{R}^{ HW\times T\times T}$.
By computing the difference between two similarity matrices, we can obtain a differential attention matrix, denoted as:
\begin{equation}
    A = A_{in} - A_{r}.
    \label{eq15}
\end{equation}
The final attention $\mathrm{Atten}_i$ is calculated inside each head as:
\begin{equation}
\mathrm{Atten}_i=\text{softmax}(A_i/\sqrt{d})V_{in,i}.
\end{equation}
Finally, the outputs of all $n_h$ attention heads are concatenated along the channel dimension, reshaped, and then processed through a linear projection as:
\begin{equation}
F_{out}=\text{Concat}_{i=1}^{n_h}\big(\mathrm{Atten}_i\big)W+b.
    \label{eq17}
\end{equation}
\subsubsection{Differential Modulated Feed-Forward Network}
To maximize the use of intermediate features, preserving useful information while discarding redundant data, we designed the DM-FFN.
DM-FFN performs pixel-level modulation on the redundant information present in the current features through feature modulation.
As shown in Fig.~\ref{fig:network} (c), DM-FFN contains two branches: a feature branch, which models local spatiotemporal information, and a modulation branch, which generates weights by leveraging both the current features and the intermediate state to modulate the enhanced features.
In the feature branch, for the input $F_m$, we first use a $1 \times 1 \times 1$ convolution to expand the channels fourfold, then apply a depth-wise convolution DConv to extract local spatiotemporal information—with a kernel size of $1 \times 3 \times 3$ in the spatial domain and $3 \times 1 \times 1$ in the temporal domain—followed by a $1 \times 1 \times 1$ convolution to reduce the dimensionality and obtain the enhanced feature representation $F_e$.
In the modulation branch, we first map the current features and the intermediate state using a $1 \times 1 \times 1$
convolution and compute their difference. The difference is refined by another $1 \times 1 \times 1$ convolution and then passed through a sigmoid function to obtain pixel-level weights, which are multiplied by the enhanced features $F_e$ to produce the modulated output $F_{mod}$. Mathematically, the process can be expressed as:
\begin{align}
F_e&=\text{ConvLayer}(F_m), \\
w &= \text{Sigmoid}(\text{Conv}_1(\text{Conv}_1(F_m)-\text{Conv}_1(S))),\\
F_{mod}&=F_e\odot w,
\end{align}
where $\text{ConvLayer}$ denotes the convolutional layer in the feature branch, $\text{Conv}_1$ denotes the $1 \times 1 \times 1$ convolution in the modulation branch, and $w$ represents the pixel-level weights obtained from the modulation branch.

\section{Experiments}
\subsection{Implementation Details}
\begin{table*}[!htbp]
  \caption{Quantitative results of different methods on grayscale videos. Params (M) $\downarrow$ and FLOPs (G) $\downarrow$ are listed per method; each per-dataset cell reports PSNR (dB) $\uparrow$, SSIM $\uparrow$. The best and second-best results are highlighted in bold and underlined, respectively.}
  \label{tab:simu}
  \centering
  \scriptsize
  \begin{tabular}{lccccccccc}
  \toprule[0.2em]
  Method
  & Params
  & FLOPs
  & Kobe
  & Traffic
  & Runner
  & Drop
  & Crash
  & Aerial
  & Average
  \\
  \midrule[0.15em]

  GAP-TV
  & -
  & -
  & {26.46, 0.885}
  & {20.89, 0.715}
  & {28.52, 0.909}
  & {34.63, 0.970}
  & {24.82, 0.838}
  & {25.05, 0.828}
  & {26.73, 0.858}
  \\

  PnP-FFDNet
  & -
  & -
  & {30.50, 0.926}
  & {24.18, 0.828}
  & {32.15, 0.933}
  & {40.70, 0.989}
  & {25.42, 0.849}
  & {25.27, 0.829}
  & {29.70, 0.892}
  \\

  DUN-3DUnet
  & 64.92
  & 4269.01
  & {35.00, 0.969}
  & {31.76, 0.966}
  & {40.03, 0.980}
  & {44.96, \textbf{0.995}}
  & {29.33, 0.956}
  & {30.46, 0.943}
  & {35.26, 0.968}
  \\

  ELP-Unfolding
  & 593.21
  & 4976.73
  & {34.41, 0.966}
  & {31.58, 0.962}
  & {41.16, 0.986}
  & {44.99, \textbf{0.995}}
  & {29.65, 0.959}
  & {30.68, 0.944}
  & {35.41, 0.969}
  \\

  STFormer
  & 20.43
  & 3286.46
  & {35.53, 0.973}
  & {32.15, 0.967}
  & {42.64, 0.988}
  & {45.08, \textbf{0.995}}
  & {31.06, 0.970}
  & {31.56, {0.953}}
  & {36.34, 0.974}
  \\

  EfficientSCI
  & 9.25
  & 1531.56
  & {35.76, 0.974}
  & {32.30, 0.968}
  & {43.05, 0.988}
  & {45.18, \textbf{0.995}}
  & {31.13, 0.971}
  & {31.50, 0.953}
  & {36.48, 0.975}
  \\

  EfficientSCI++
  & 8.47
  & \underline{1141.40}
  & {35.70, 0.974}
  & {32.24, 0.968}
  & {42.73, {0.988}}
  & {45.33, \textbf{0.995}}
  & {31.02, 0.971}
  & {31.64, 0.953}
  & {36.44, 0.975}
  \\

  HiSViT
  & 9.42
  & 1649.18
  & {\underline{36.24}, \underline{0.976}}
  & {33.06, \underline{0.973}}
  & {43.84, \textbf{0.991}}
  & {45.55, \textbf{0.995}}
  & {31.62, \underline{0.976}}
  & {31.67, 0.957}
  & {37.00, \underline{0.978}}
  \\

  DADUN
  & 13.29
  & 6728.23
  & {36.23, \underline{0.976}}
  & {\underline{33.08}, 0.972}
  & {43.60, 0.990}
  & {45.49, \textbf{0.995}}
  & {\underline{31.93}, \underline{0.976}}
  & {32.00, 0.958}
  & {37.06, \underline{0.978}}
  \\
    \midrule[0.15em]

  DU-5stg (Ours)
  & \textbf{4.14}
  & \textbf{965.99}
  & {{36.11}, \underline{0.976}}
  & {{32.80}, {0.971}}
  & {\underline{43.86}, 0.990}
  & {\underline{45.84}, \textbf{0.995}}
  & {31.79, 0.975}
  & {\underline{32.18}, \underline{0.959}}
  & {\underline{37.10}, \underline{0.978}}
  \\

  DU-9stg  (Ours)
  & \underline{6.67}
  & 1690.47
  & {\textbf{36.51}, \textbf{0.978}}
  & {\textbf{33.45}, \textbf{0.974}}
  & {\textbf{44.75}, \textbf{0.991}}
  & {\textbf{45.99}, \textbf{0.995}}
  & {\textbf{32.47}, \textbf{0.978}}
  & {\textbf{32.27}, \textbf{0.961}}
  & {\textbf{37.57}, \textbf{0.980}}
  \\
  \bottomrule[0.1em]
  \end{tabular}
\end{table*}

\begin{figure*}[t]
   \centering
    \includegraphics[width=0.95\linewidth]{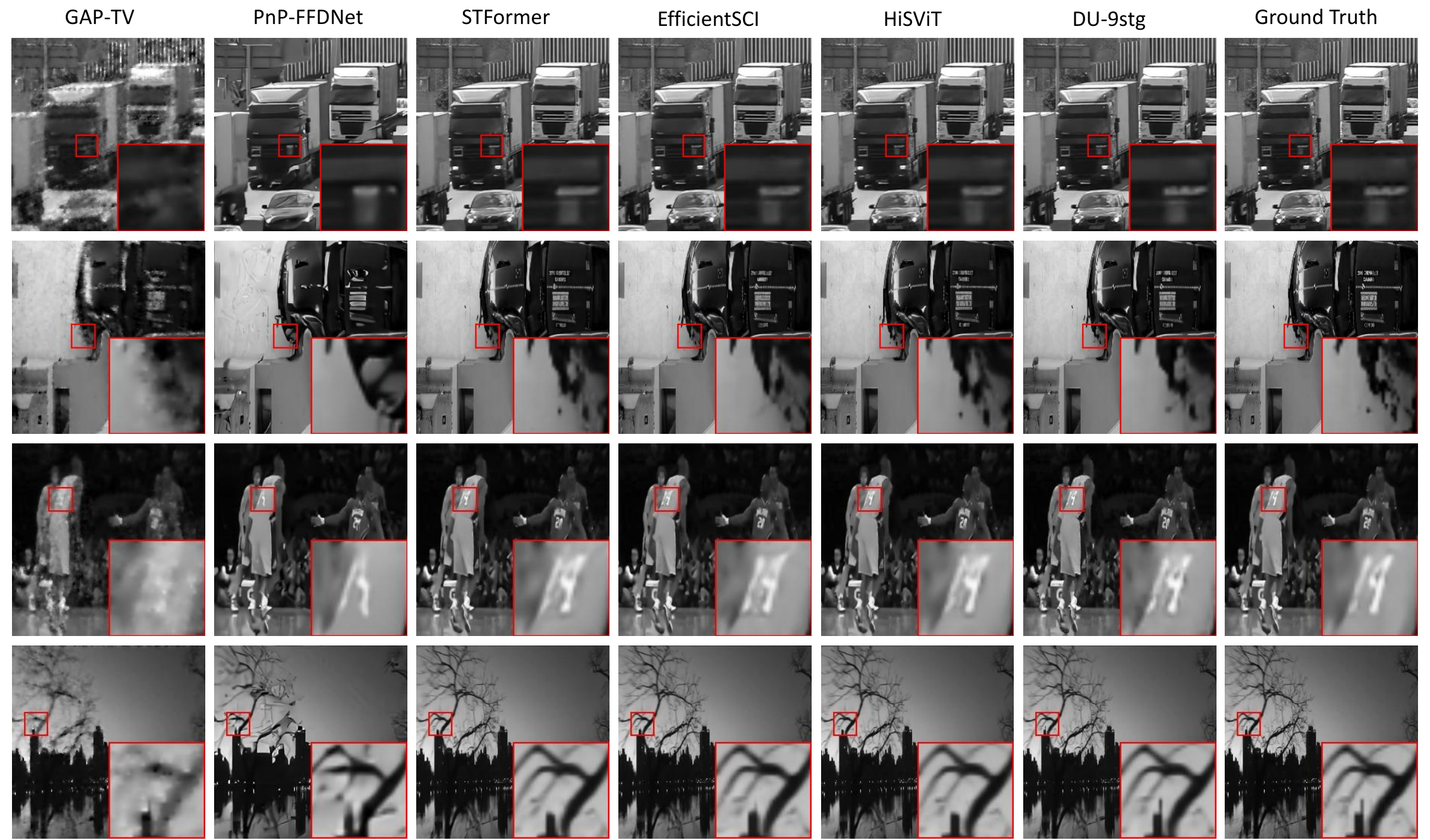}
    \caption{Comparison of reconstruction results of different methods (GAP-TV~\cite{gap-tv}, PnP-FFDNet~\cite{pnp-ffdnet}, STFormer~\cite{stformer}, EfficientSCI~\cite{efficientsci}, HiSViT~\cite{hisvit}, DU-9stg (Ours)) on representative scenes from the Traffic, Crash, Kobe, and Aerial grayscale benchmark datasets. Zoom in for a better view.}
 \label{fig:gray_sim}

\end{figure*}
\subsubsection{Experiment Setting}
The proposed method is implemented using PyTorch and trained on a single RTX PRO 6000 GPU.
Following previous works, our models are trained in the DAVIS2017 dataset~\cite{davis} with the same data augmentation in \cite{efficientsci}.
To verify model performance, we conduct experiments on six grayscale/color benchmark videos with a resolution of $8\times256\times256$ / $8\times512\times512\times3$ pixels and on real captured grayscale videos (Domino and Water Balloon~\cite{e2e-cnn}) with the resolution of $10\times512\times512$ pixels.
We adopt the Adam optimizer~\cite{adam} with settings $\beta_1=0.9$ and $\beta_2=0.999$, and the initial learning rate is set to $4\times10^{-4}$ and adopt the Cosine Annealing learning rate scheme ~\cite{sgdr} to implement end-to-end training.
After training for over 100 epochs on the pixels of $8\times128\times128 (\times3)$, we continued training for over 30 epochs on the pixels of $8\times256\times256 (\times3)$. For the real captured data, which compress 10 frames into one measurement, the model is retrained under the corresponding 10-frame setting.
The Peak Signal-to-Noise Ratio (PSNR) and Structural Similarity Index Metric (SSIM) \cite{psnr} are used as performance metrics to evaluate the reconstruction quality. 
To supervise intermediate stages, we adopt a full-stage loss with exponential per-stage weighting,
\begin{equation}
L=\frac{1}{N}\sum_{i=1}^{N}\sum_{m=1}^{M}e^{\beta m}\big\|\bar{X}_{m}^{i}-X^{i}\big\|_2^2,
\end{equation}
where $N$ is the number of training samples, $M$ the number of unfolding stages, $\bar{X}_{m}^{i}$ the reconstruction of the $m$-th stage, and $X^{i}$ the ground truth. A larger $\beta$ places more weight on later stages, which are closer to the final output; we set $\beta=0.1$. Our ablation shows that supervising all stages consistently outperforms supervising only the final stage, that a moderate $\beta$ works best, and that an overly large $\beta$ over-emphasizes late stages and degrades performance; the detailed results are provided in the supplementary material.

\subsection{Results on Grayscale Simulation Videos}
We selected previous state-of-the-art (SOTA) models for comparison, including model-based methods (GAP-TV~\cite{gap-tv}), plug-and-play methods (PnP-FFDNet~\cite{pnp-ffdnet}),
Transformer-based methods (STFormer~\cite{stformer}, EfficientSCI~\cite{efficientsci}, EfficientSCI++~\cite{efficientsci++}, HiSViT~\cite{hisvit}) and deep unfolding methods (DUN-3DUnet~\cite{dun-3dunet}, ELP-Unfolding~\cite{elp}, DADUN~\cite{dadun}) on 6 simulated grayscale datasets.
To intuitively show the effectiveness of our DU, we provide Performance-FLOPs-Params comparisons of SOTA methods in Fig.~\ref{fig:psnr}.
Table~\ref{tab:simu} presents the detailed quantitative comparison results of our method with other methods in terms of PSNR, SSIM, parameters, and FLOPs.
It can be observed that DU-5stg achieves the highest average PSNR among the compared prior methods (37.10 dB versus 37.06 dB for DADUN) with the fewest parameters and FLOPs, although HiSViT or DADUN remain slightly better on individual scenes such as Kobe, Traffic, and Crash. DU-9stg further raises the average PSNR to 37.57 dB.
Compared with the previous method with the lowest parameter count and computational cost, EfficientSCI++, our method achieves an improvement of over 0.6 dB.
Compared to the previous best deep unfolding method DADUN, our DU-9stg only requires 50.2\% of the parameters and 25.12\% of the FLOPs while achieving an improvement of 0.51 dB.
DU-9stg also obtains the best PSNR on all six scenes, and surpasses the strongest Transformer-based method HiSViT by 0.57 dB in average PSNR with 29\% fewer parameters at a comparable computational cost.
Fig.~\ref{fig:gray_sim} presents a visual comparison between DU and several other methods. By magnifying certain local regions, our DU demonstrates the ability to reconstruct sharper edges and finer details.

\subsection{Results on Color Simulation Videos}
\begin{table*}[!t]
  \caption{Quantitative results of different methods on color videos. Params (M) $\downarrow$ and FLOPs (G) $\downarrow$ are listed per method; each per-dataset cell reports PSNR (dB) $\uparrow$, SSIM $\uparrow$. The best and second-best
results are highlighted in bold and underlined, respectively.}
  \label{tab:color}
  \centering
  \scriptsize
  \resizebox{\textwidth}{!}{
  \begin{tabular}{lccccccccc}
  \toprule[0.2em]
  Method
  & Params
  & FLOPs
  & Beauty
  & Bosphorus
  & Jockey
  & Runner
  & ShakeNDry
  & Traffic
  & Average
  \\
  \midrule[0.15em]

  GAP-TV
  & -
  & -
  & \makecell[c]{33.08,  0.964}
  & \makecell[c]{29.70,  0.914}
  & \makecell[c]{29.48,  0.887}
  & \makecell[c]{29.10,  0.878}
  & \makecell[c]{29.59,  0.893}
  & \makecell[c]{19.84,  0.645}
  & \makecell[c]{28.47,  0.864}
  \\

PnP-FFDNet
  & -
  & -
  & \makecell[c]{34.15,  0.967}
  & \makecell[c]{33.06,  0.957}
  & \makecell[c]{34.80,  0.943}
  & \makecell[c]{35.32,  0.940}
  & \makecell[c]{32.37,  0.940}
  & \makecell[c]{24.55,  0.837}
  & \makecell[c]{32.38,  0.931}
  \\

BIRNAT
  & \textbf{4.34}
  & \textbf{1562.25}
  & \makecell[c]{36.08,  0.975}
  & \makecell[c]{38.30,  0.982}
  & \makecell[c]{36.51,  0.956}
  & \makecell[c]{39.65,  0.973}
  & \makecell[c]{34.26,  0.951}
  & \makecell[c]{28.03,  0.915}
  & \makecell[c]{35.47,  0.959}
  \\

  STFormer
  & 20.44
  & 13051.84
  & \makecell[c]{37.37,  \textbf{0.981}}
  & \makecell[c]{40.39,  0.988}
  & \makecell[c]{38.32,  0.968}
  & \makecell[c]{42.45,  {0.985}}
  & \makecell[c]{35.15,  0.956}
  & \makecell[c]{{30.24},  {0.939}}
  & \makecell[c]{37.32,  {0.970}}
  \\

  EfficientSCI
  & 9.25
  & {6121.94}
  & \makecell[c]{{37.51},  {0.979}}
  & \makecell[c]{{40.89},  {0.988}}
  & \makecell[c]{{38.49},  {0.969}}
  & \makecell[c]{42.73,  {0.985}}
  & \makecell[c]{{35.19},  {0.953}}
  & \makecell[c]{{30.13},  {0.943}}
  & \makecell[c]{{37.49},  {0.970}}
  \\

EfficientSCI++
  & {8.47}
  & \underline{4570.06}
  & \makecell[c]{{37.53},  {0.980}}
  & \makecell[c]{{40.96},  {0.989}}
  & \makecell[c]{{38.50},  {0.969}}
  & \makecell[c]{{42.75},  {0.985}}
  & \makecell[c]{35.18,  {0.953}}
  & \makecell[c]{{30.09},  {0.942}}
  & \makecell[c]{{37.50},  {0.970}}
  \\
DADUN
  & 10.95
  & 21916.26
  & \makecell[c]{37.48, \textbf{0.981}}
  & \makecell[c]{41.49, \textbf{0.990}}
  & \makecell[c]{\textbf{39.37}, \textbf{0.973}}
  & \makecell[c]{\underline{43.34}, \underline{0.986}}
  & \makecell[c]{35.32, 0.957}
  & \makecell[c]{30.31, 0.940}
  & \makecell[c]{37.89, 0.972}
  \\
    HiSViT
  & 9.42
  & 6596.73
  & \makecell[c]{\underline{37.75},\textbf{ 0.981}}
  & \makecell[c]{\underline{41.50}, \textbf{0.990}}
  & \makecell[c]{\underline{39.29}, \underline{0.972}}
  & \makecell[c]{43.27, \underline{0.986}}
  & \makecell[c]{\underline{35.49}, \underline{0.958}}
  & \makecell[c]{\underline{30.79}, \underline{0.946}}
  & \makecell[c]{\underline{38.01}, 0.972}
  \\
    \midrule[0.15em]
DU-9stg (Ours)
  & \underline{6.67}
  & 6763.09
  & \makecell[c]{\textbf{37.92}, \textbf {0.981}}
  & \makecell[c]{\textbf{41.76}, \textbf{0.990}}
  & \makecell[c]{39.28, \underline{0.972}}
  & \makecell[c]{\textbf{43.59}, \textbf{0.987}}
  & \makecell[c]{\textbf{35.73}, \textbf{0.960}}
  & \makecell[c]{{\textbf{31.53}}, \textbf{0.952}}
  & \makecell[c]{\textbf{38.30}, \textbf{0.974}}
  \\
  \bottomrule[0.1em]
  \end{tabular}
  }
\end{table*}
\begin{figure*}[!t]
    \centering
    \includegraphics[width=0.95\linewidth]{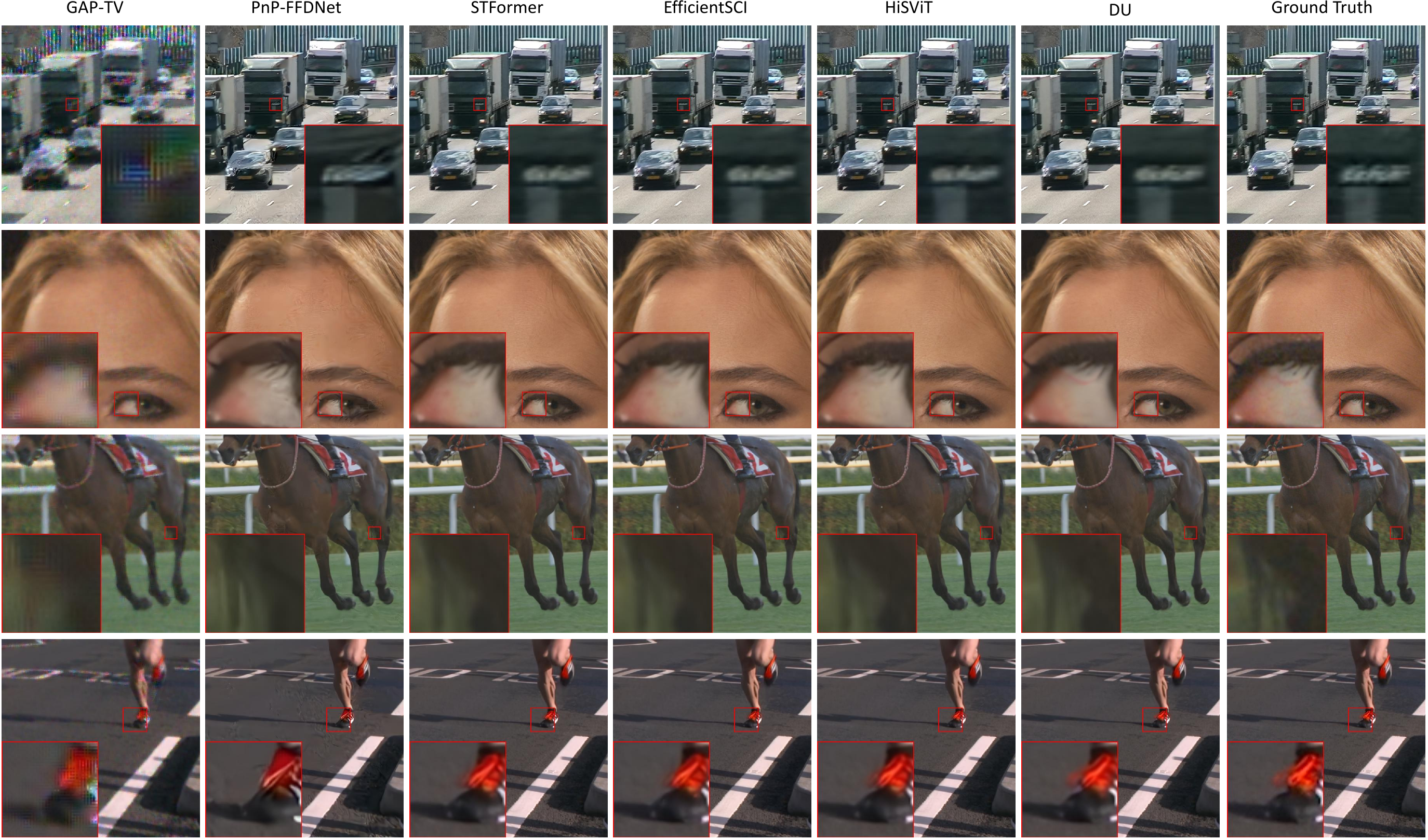}
    \caption{Comparison of reconstruction results of different methods (GAP-TV~\cite{gap-tv}, PnP-FFDNet~\cite{pnp-ffdnet}, STFormer~\cite{stformer}, EfficientSCI~\cite{efficientsci}, HiSViT~\cite{hisvit}, DU-9stg (Ours)) on representative scenes from the Traffic, Beauty, and Jockey color benchmark datasets. Zoom in for a better view.}
    \label{fig:color_sim}
\end{figure*}
For color simulation videos,  we selected previous state-of-the-art (SOTA) models for comparison, including model-based methods (GAP-TV~\cite{gap-tv}), plug-and-play methods (PnP-FFDNet~\cite{pnp-ffdnet}), RNN-based method BIRNAT~\cite{birnat}, Transformer-based methods (STFormer~\cite{stformer}, EfficientSCI~\cite{efficientsci}, EfficientSCI++~\cite{efficientsci++}, HiSViT~\cite{hisvit}), and deep unfolding method DADUN~\cite{dadun} for comparison.
Table~\ref{tab:color}
presents the detailed quantitative comparison results of our method with other methods in terms of PSNR, SSIM, parameters, and FLOPs.
Compared with the Transformer-based HiSViT, DU-9stg achieves a higher average PSNR (38.30 dB versus 38.01 dB) with fewer parameters and comparable cost, although HiSViT and DADUN remain marginally better on a few scenes such as Jockey.
Fig.~\ref{fig:color_sim} presents a visual comparison between DU and several other methods. By magnifying certain local regions, our DU demonstrates the ability to reconstruct sharper edges and finer details.

\subsection{Results on Large-Scale Color Simulation Video}
\label{sec:results_color_1080p}
\begin{table*}[!t]
  \renewcommand{\arraystretch}{1.0}
  \centering
  \caption{Quantitative results of different methods on 6 large-scale color benchmark datasets. Each cell reports PSNR (dB) $\uparrow$, SSIM $\uparrow$. Best results are in bold.}
  \label{tab:1080p}
  \scriptsize
  \resizebox{\textwidth}{!}{
  \begin{tabular}{lccccccc}
  \toprule[0.2em]
  Method
  & Jockey
  & ShakeNDry
  & Beauty
  & YachtRide
  & HoneyBee
  & ReadySteadyGo
  & Average
  \\
  \midrule[0.15em]
  GAP-TV
  & {29.83, 0.929}
  & {30.16, 0.919}
  & {33.27, 0.859}
  & {26.66, 0.887}
  & {34.11, 0.972}
  & {25.45, 0.805}
  & {29.91, 0.895}
  \\
  PnP-FastDVDnet
  & {36.10, 0.970}
  & {35.30, 0.964}
  & {35.03, 0.877}
  & {34.76, 0.973}
  & {39.90, 0.989}
  & {35.10, 0.957}
  & {36.03, 0.942}
  \\
  STFormer
  & {38.02, 0.978}
  & {\textbf{36.90}, \textbf{0.973}}
  & {36.13, 0.894}
  & {37.03, 0.982}
  & {40.75, 0.991}
  & {37.92, 0.973}
  & {37.79, 0.965}
  \\
  EfficientSCI
  & {37.63, 0.976}
  & {36.67, 0.971}
  & {35.93, 0.891}
  & {36.81, 0.981}
  & {40.23, 0.991}
  & {37.67, 0.953}
  & {37.49, 0.963}
  \\
  \midrule[0.15em]
  DU-9stg (Ours)
  & {\textbf{38.87}, \textbf{0.981}}
  & {36.86, \textbf{0.973}}
  & {\textbf{36.17}, \textbf{0.895}}
  & {\textbf{37.36}, \textbf{0.984}}
  & {\textbf{41.11}, \textbf{0.992}}
  & {\textbf{39.15}, \textbf{0.978}}
  & {\textbf{38.25}, \textbf{0.967}}
  \\
  \bottomrule[0.1em]
  \end{tabular}
  }
\end{table*}
\begin{figure*}[!t]
    \centering
    \includegraphics[width=.98\textwidth]{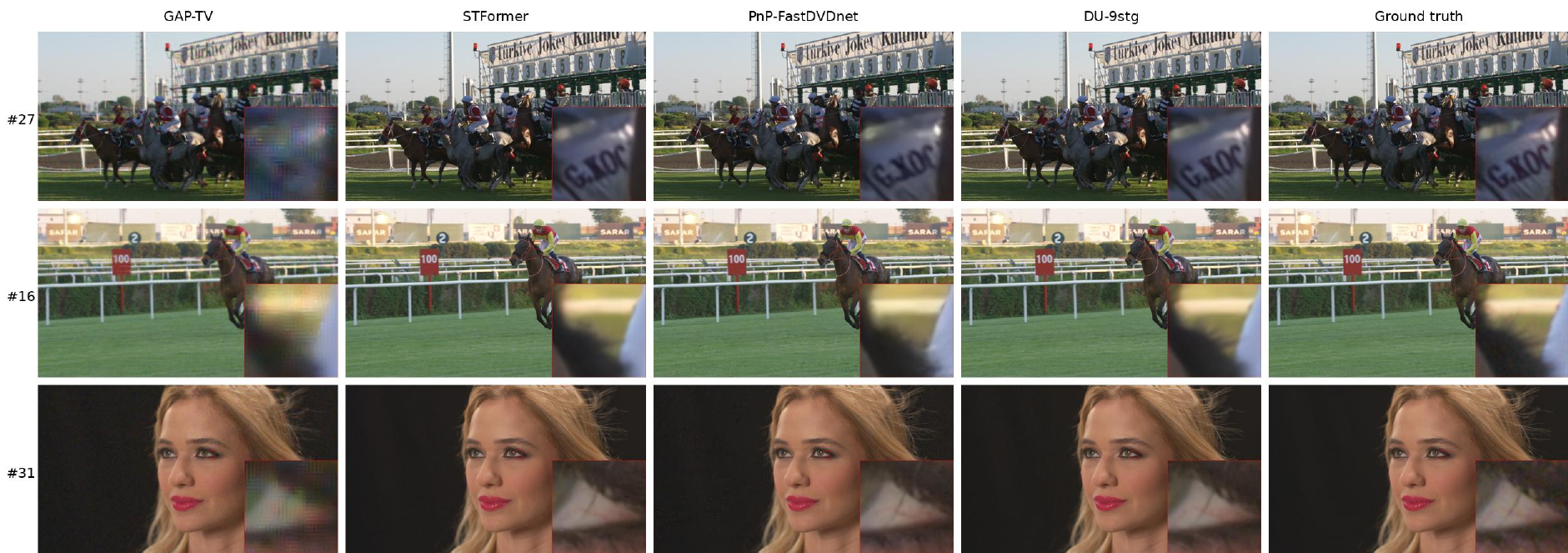}
    \caption{Comparison of reconstruction results of different methods (GAP-TV~\cite{gap-tv}, STFormer~\cite{stformer}, PnP-FastDVDnet~\cite{pnp-fastdcdnet}, DU-9stg (Ours)) on large-scale $1920\times1080$ UVG benchmark scenes~\cite{uvg}: ReadySteadyGo (frame \#27), Jockey (frame \#16), and Beauty (frame \#31). Zoom in for a better view.}
  \label{fig:1080p}
\end{figure*}

Due to the high memory demand of large-resolution reconstruction, memory-intensive methods such as HiSViT~\cite{hisvit} cannot be run at larger resolutions. To evaluate DU on high-resolution data, we selected 6 datasets of size $1920\times1080\times8\times3$ from the Ultra Video Group (UVG) dataset~\cite{uvg}, and compared against model-based methods (GAP-TV~\cite{gap-tv}, PnP-FastDVDnet-color~\cite{pnp-fastdcdnet}) and end-to-end methods (STFormer~\cite{stformer}, EfficientSCI~\cite{efficientsci}). Table~\ref{tab:1080p} reports the PSNR/SSIM comparison. In terms of reconstruction fidelity, DU-9stg outperforms previous model-based and plug-and-play methods by a large margin ($>$2 dB), and outperforms STFormer by 0.46 dB. Fig.~\ref{fig:1080p} shows the corresponding visual comparison; DU retrieves more details and textures than competing methods.

\subsection{Results on Real Captured Video}
\begin{figure*}[!t]
    \centering
    \includegraphics[width=0.95\linewidth]{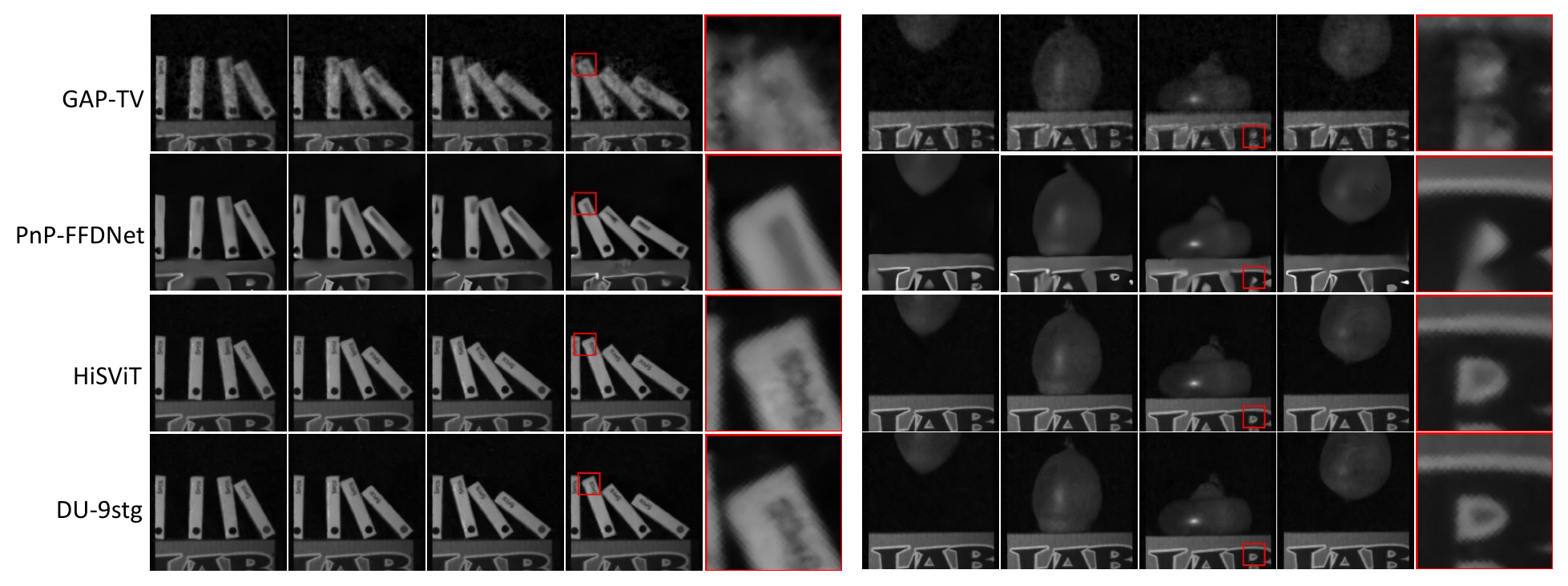}
    \caption{Comparison of reconstruction results of different methods (GAP-TV~\cite{gap-tv}, PnP-FFDNet~\cite{pnp-ffdnet}, HiSViT~\cite{hisvit}, DU-9stg (Ours)) on real captured Domino and Water Balloon video data~\cite{e2e-cnn}. Zoom in for a better view.}
    \label{fig:real}
\end{figure*}

We provide the results of DU compared with GAP-TV~\cite{gap-tv}, PnP-FFDNet~\cite{pnp-ffdnet}, and HiSViT~\cite{hisvit} on real datasets. Fig.~\ref{fig:real} shows the visual reconstruction results.
We magnify local regions to obtain a clearer comparison. Our method can reconstruct sharper textures and detailed features. Specifically, it achieves the clearest reconstruction of the letters on the dominoes and Water Balloon.
\subsection{Ablation Study}
To evaluate the contribution of each component in the DU, as well as the impact of different periodicity designs of DEF and the influence of DRP on other unfolding methods, we performed ablation experiments on grayscale videos.

\begin{table}[!tb]
\centering
\caption{Effect of DRP on other unfolding methods. * means that we apply the DRP to the corresponding method, with the periodicity set to 4.}
\label{tab:ablation2}
\footnotesize
\begin{tabular}{lcccc}
\toprule[0.2em]
\textbf{Method} & \textbf{Params (M)} & \textbf{FLOPs (G)}  & \textbf{PSNR (dB)}   & \textbf{SSIM}  \\
\midrule[0.15em]
DUN-3DUnet &  64.92  & 4269.01 & 30.75  & 0.9188\\
DUN-3DUnet* & 16.59 & 1354.61 & 30.91 & 0.9193\\
DADUN & 7.46 & 3603.70 & 33.28 & 0.9495\\
DADUN* & 4.24 & 1880.47 & 33.58 & 0.9522\\
\bottomrule[0.1em]
\end{tabular}
\end{table}

\subsubsection{Periodicity of DEF}
\begin{figure}[!t]
\centering
\includegraphics[width=0.95\columnwidth]{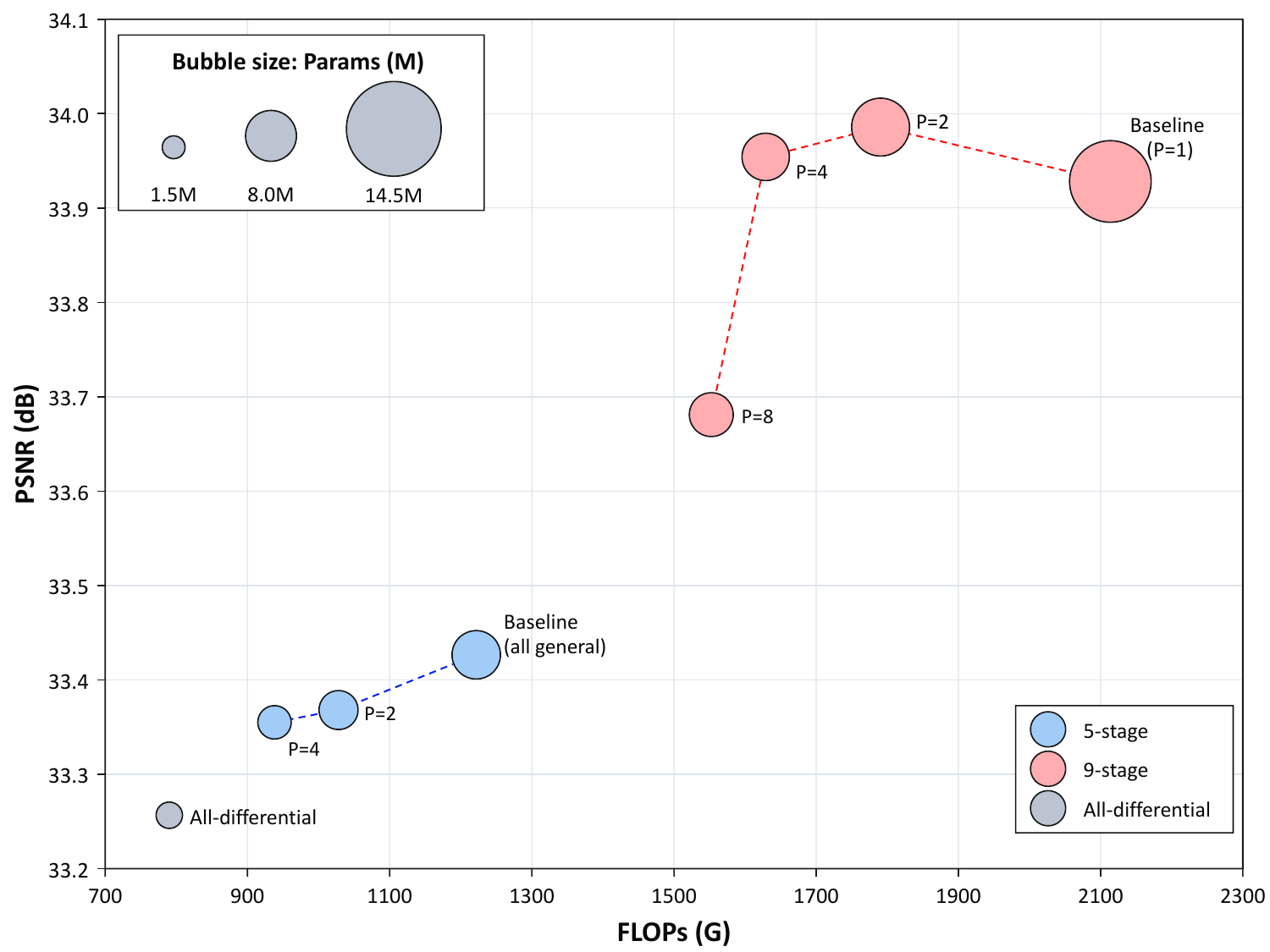}
\caption{Trade-off between reconstruction quality and computational cost under different unfolding periodicities. The horizontal axis denotes FLOPs (G), the vertical axis denotes PSNR (dB), and bubble size represents the number of parameters (M). Blue and red bubbles denote the 5-stage and 9-stage models, respectively, while gray denotes the all-differential variant.}
\label{fig:periodicity}
\end{figure}

To investigate the impact of different periodicities $P$ on the reconstruction results, we set the model without a differential stage (i.e., $P=1$) as Baseline, and conduct ablation experiments on models with 5 and 9 stages for various $P$.
As shown in Fig.~\ref{fig:periodicity}, $P=4$ achieves a favorable balance between accuracy and cost. Compared with the uniformly general baseline ($P=1$), it reduces parameters by 48\% and 54\% and FLOPs by 20\% and 22\% for the 5-stage and 9-stage models, respectively, while changing PSNR by only $-0.06$~dB and $+0.03$~dB. Conversely, the all-differential variant 
is the lightest but trails $P=4$ by 0.11~dB in the 5-stage setting, indicating that the periodic general stages act as useful anchors.

\subsubsection{DRP for other unfolding methods}
To evaluate the effectiveness of DRP in other unfolding methods, we applied DRP to DUN-3DUnet and DADUN for ablation experiments, with the periodicity set to 4. As shown in the Table~\ref{tab:ablation2}, applying DRP to the aforementioned unfolding methods substantially reduces both computational cost and parameter count, while also improving performance, demonstrating that our DRP can be broadly applied to other unfolding frameworks to achieve superior results.

\subsubsection{Attention Comparison}
To study the effect of our DRA, we perform ablation with other attention designs.
Baseline-2 is obtained by removing DRA from DU-5stg.
G-MSA~\cite{g-msa} and F-MSA~\cite{vivit} are impractical because their computational complexities scale quadratically with the spatial-temporal resolution $T \times H \times W$ and the spatial resolution $ H \times W$, respectively.
BDA~\cite{ctm-sci} computes spatiotemporal correlations simultaneously within a 3D window.
SLTW-MSA~\cite{stformer} adopts a space-time factorization mechanism and computes spatial-window attention and temporal attention separately.
CSS-MSA~\cite{hisvit} attends to all spatial-temporal tokens with cross-scale interactions in a single attention layer.
As shown in Table~\ref{tab:ablation3}, baseline-2 yields 32.88 dB.
DRA yields the largest improvement among the compared attention designs.

\begin{table}[t]
\centering
\caption{Comparison of DRA and competitive Attentions.}
\label{tab:ablation3}
\footnotesize
\begin{tabular}{lcccc}
\toprule[0.2em]
\textbf{Method} & \textbf{Params (M)} & \textbf{FLOPs (G)}  & \textbf{PSNR (dB)}   & \textbf{SSIM} \\
\midrule[0.15em]
Baseline-2 & 3.95 & 868.74 & 32.88 & 0.9463 \\
BDA & 4.08 & 937.16 & 33.00 & 0.9466 \\
SLTW-MSA & 4.08 & 937.09 & 33.15 & 0.9484\\
CSS-MSA & 4.09 & 936.99 & 33.22 & 0.9495 \\
DRA & 4.14 & 965.99 & 33.37& 0.9502\\
\bottomrule[0.1em]
\end{tabular}
\end{table}

\subsubsection{Feed-Forward Network Comparison}
To study the effect of our DM-FFN, we perform ablation with other FFN designs.
Baseline-3 is obtained by removing DM-FFN from DU-5stg.
GR-FFN~\cite{stformer} conducts more feature mapping by grouping and using the residual mechanism to achieve multi-layer information fusion.
GSM-FFN~\cite{hisvit} strengthens the locality by introducing gated self-modulation to regular FFN.
As shown in Table~\ref{tab:ablation4}, baseline-3 yields 32.85 dB.
DM-FFN yields the largest improvement among the compared FFN designs.

\begin{table}[!tb]
\centering
\caption{Comparison of DM-FFN and competitive FFN.}
\label{tab:ablation4}
\footnotesize
\begin{tabular}{lcccc}
\toprule[0.2em]
\textbf{Method} & \textbf{Params (M)} & \textbf{FLOPs (G)}  & \textbf{PSNR (dB)}   & \textbf{SSIM}  \\
\midrule[0.15em]
Baseline-3 &  3.77  & 776.34 & 32.85  & 0.9461\\
Regular FFN & 4.06 & 926.12 & 32.98 & 0.9465\\
GR-FFN & 4.53 & 1168.92 & 32.57 & 0.9455\\
GSM-FFN & 4.45 & 1125.43 & 33.27 & 0.9497\\
DM-FFN & 4.14 & 965.99 &33.37 &0.9502\\
\bottomrule[0.1em]
\end{tabular}
\end{table}

\subsubsection{Model Flexibility for Varying Masks}
We evaluate the robustness of DU-9stg to sensing matrix variation by testing it, along with
EfficientSCI~\cite{efficientsci} and HiSViT~\cite{hisvit}, on five newly generated random
binary masks that differ from the mask used during training, in addition to the pre-trained
mask. As reported in Table~\ref{tab:ablation_mask}, mask changes slightly affect the
performance of all pre-trained models. DU-9stg shows the smallest average PSNR change
among the three methods, and all three remain within 0.02~dB of their pre-trained results,
indicating that DU is at least as robust to sensing matrix variation as the compared methods.

\begin{table}[!tb]
\centering
\caption{Robustness evaluation under different masks (PSNR).}
\label{tab:ablation_mask}
\footnotesize
\setlength{\tabcolsep}{3pt}
\resizebox{\columnwidth}{!}{
\begin{tabular}{l c c c c c c c}
\toprule[0.2em]
\textbf{Method} & \textbf{Pretr.} & \textbf{R1} & \textbf{R2} & \textbf{R3} & \textbf{R4} & \textbf{R5} & \textbf{Avg. Var.} \\
\midrule[0.15em]
EfficientSCI & 36.3948 & 36.3619 & 36.3938 & 36.3214 & 36.4208 & 36.3794 & -0.0193 \\
HiSViT & 36.9953 & 36.9985 & 36.9955 & 36.9513 & 36.9771 & 36.9610 & -0.0186 \\
DU-9stg (Ours) & 37.5740 & 37.5527 & 37.5418 & 37.5735 & 37.5600 & 37.5639 & \textbf{-0.0156} \\
\bottomrule[0.1em]
\end{tabular}
}
\end{table}

\section{Conclusion}
We proposed Differential Unfolding (DU), which alternates sparse high-capacity general stages with lightweight differential stages built on DRP, achieving state-of-the-art video SCI reconstruction at a much lower cost than prior unfolding methods. A limitation is the extra inference overhead of attention in DRA, which we plan to reduce with more efficient attention, such as linear attention~\cite{la}.

\bibliographystyle{IEEEtran}
\bibliography{IEEEabrv,main}

\end{document}